\documentclass[sigconf]{acmart}

\usepackage{multirow}
\usepackage{multicol}
\usepackage{booktabs}
\usepackage{epsfig}

\usepackage{url}
\usepackage{array}
\usepackage{color}

\graphicspath{ {./images/} }
\usepackage{subcaption}

\copyrightyear{2020} 
\acmYear{2020} 
\setcopyright{acmcopyright}\acmConference[MM '20]{Proceedings of the 28th ACM International Conference on Multimedia}{October 12--16, 2020}{Seattle, WA, USA}
\acmBooktitle{Proceedings of the 28th ACM International Conference on Multimedia (MM '20), October 12--16, 2020, Seattle, WA, USA}
\acmPrice{15.00}
\acmDOI{10.1145/3394171.3416300}
\acmISBN{978-1-4503-7988-5/20/10}

\begin{document}

\title{A Simple Baseline for Pose Tracking in Videos of Crowded Scenes}

\author{Li Yuan}
\authornote{Authors contributed equally to this work; Work done during internship at YITU Technology.}
\email{yuanli@u.nus.edu}
\orcid{0000-0002-2120-5588}
\author{Shuning Chang}
\authornotemark[1]
\author{Ziyuan Huang}
\authornotemark[1]
\affiliation{
  \institution{National University of Singapore}
}

\author{Yichen Zhou}
\affiliation{%
  \institution{YITU Technology}
  \institution{National University of Singapore}
}

\author{Yunpeng Chen}
\affiliation{%
  \institution{YITU Technology}
}

\author{Xuecheng Nie}
\affiliation{%
 \institution{YITU Technology}}

\author{Francis E.H. Tay}
\affiliation{%
  \institution{National University of Singapore}
  }

\author{Jiashi Feng}
\affiliation{%
  \institution{National University of Singapore}
  }

\author{Shuicheng Yan}
\affiliation{%
  \institution{YITU Technology}}

\renewcommand{\shortauthors}{Yuan and Huang, et al.}

\begin{abstract}
  This paper presents our solution to ACM MM challenge: Large-scale Human-centric Video Analysis in Complex Events~\cite{lin2020human}; specifically, here we focus on Track3: Crowd Pose Tracking in Complex Events. Remarkable progress has been made in multi-pose training in recent years. However, how to track the human pose in crowded and complex environments has not been well addressed. We formulate the problem as several subproblems to be solved. First, we use a multi-object tracking method to assign human ID to each bounding box generated by the detection model. After that, a pose is generated to each bounding box with ID. At last, optical flow is used to take advantage of the temporal information in the videos and generate the final pose tracking result.
\end{abstract}



\maketitle

\section{Introduction}
\begin{figure*}
    \centering
    \includegraphics[width=1\textwidth]{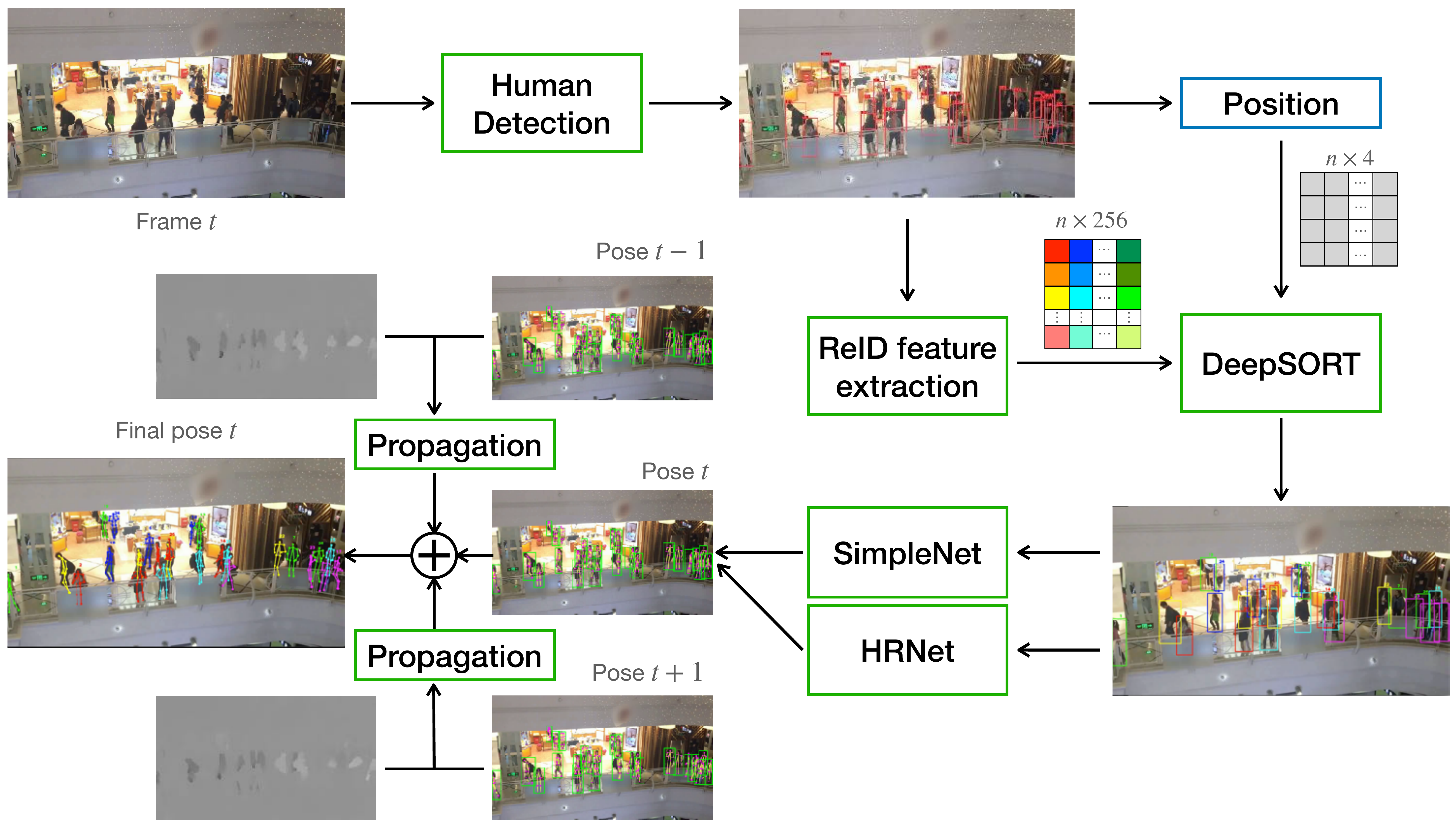}
    \caption{The overall structure of the proposed solution to human pose tracking. The solution is formulated as two stages. In the first stage, humans in every video frames are detected and DeepSORT is applied to associate boxes in different frames. For DeepSORT tracking, we used our own pretrained ReID model. Combined with position information, DeepSORT is able to perform well on the test set of HIE. Second stage involves human pose estimation as well as tracking. The pose estimation is performed on each box in the tracking result. Taking advantage of the pose in the last frame as well as the subsequent frame, the pose of the current frame is smoothed temporally. The identity of the box is automatically transferred to the final poses with the corresponding bounding boxes.}
    \label{fig:structure}
\end{figure*}

The general pipeline for the pose tracking method that we used can be divided into two parts, respectively human tracking and pose estimation. First, we use a multi-object tracking method to assign an id to each human boxes generated by the detection phase. During human tracking, some detection boxes will be refined or deleted. In the second step, we perform pose estimation on every box that is assigned an id during tracking. The result of the pose estimation combined with the id assigned by the tracking method is the final submission result. The overall structure of our method can be seen in Fig.~\ref{fig:structure}.

Since the problem is treated as a two-stage problem to be tackled one by one, each module will be introduced separately. The following of the report is organized as follows: Sec.~\ref{sec:human_det} introduces the human detection model that we used; Sec.~\ref{sec:human_tracking} introduces the human tracking algorithm; Sec.~\ref{sec:pose_estimation} introduces the pose estimation as well as the final pose generation process. Finally, Sec.~\ref{sec:conclusion} concludes the report.

\section{Human Detection}
\label{sec:human_det}
The first step of human pose tracking is to detect the bounding boxes of person. As no validation set in HIE dataset~\cite{lin2020human}, we split the original training set as a new training set and validation set. Two splitting strategies are tried: splitting by image frames (5k for validation, 27k for training) and splitting by videos. We found that splitting by image will cause over-fitting and far away from the data distribution of the testing set. So we adopt the video-splitting strategy and split video 3,7,8 and 17 as the validation set and the rest videos as train data, in which 5.7k image frames for validation and the reset 27k images frames for training. Based on the train and validation set, we can conduct detection experiments on HIE. All the performance of models is tested by two metrics, Averaged Precision (AP) and MMR~\cite{dollar2011pedestrian}. AP reflects both the precision and recall ratios of the detection results; MMR is the log-average Miss Rate on False Positive Per Image (FPPI) in $[0.01,100]$, is commonly used in pedestrian detection. MR is very sensitive to false positives (FPs), especially FPs with high confidences will significantly harm the MMR ratio. Larger AP and smaller MMR indicates better performance.

\subsection{Detection Smoothing by Optical Flow}
For human tracking in video, smoothing bounding boxes on human instances can be beneficial to reduce the tracking miss rate. We propose to smooth the current bounding box of human instance based on the previous frame by optical flow. Give one detected bounding box with joints coordinates set $I_{i}^{k}$ (including two points: ($x_1$, $y_1$) for upper right point and ($x_2$, $y_2$) as lower left point) in current frame $J^k$, we compute the $I_{i}^{k-1}$ in frame $J^{k-1}$, and the optical flow field $F^{k-1\xrightarrow{}k}$ between the two consequent frames. Then the prediction joints coordinates set $\hat{I}_i^{k}$ can be estimated by the optical flow field $F^{k-1\xrightarrow{}k}$ and the detection coordinate $I_{i}^{k-1}$ in last frame, i.e. $\hat{I}_i^{k} = I_{i}^{k-1} + F^{k-1\xrightarrow{}k}$. Similar with the Kalman filter, the current position is the weight sum of prediction and 
observations, thus we have the ``real'' bounding box in current frame is weight sum of detected box and estimated box: $\tilde{I}_i^k = \alpha \hat{I}_i^{k} + (1-\alpha)I_{i}^{k}$, where $\alpha$ is a weight and it is 0.5 in most of cases. 

\subsection{Common Methods on HIE}
There are mainly two different types of common detection frameworks: one-stage (unified) frameworks~\cite{redmon2016you, redmon2016yolo9000, liu2016ssd} and two-stage (region-based) framework~\cite{girshick2014rich, girshick2015fast, ren2015faster, he2017mask}. Since RCNN~\cite{girshick2014rich} has been proposed, the two-stage detection methods have been widely adopted or modified~\cite{ren2015faster, lin2017focal, lin2017FPN, cai2018cascade, zhou2018object, wang2019few, wang2019distilling}. Normally, the one-stage frameworks can run in real-time but with the cost of a drop in accuracy compared with two-stage frameworks, so we mainly adopt two-stage frameworks on HIE dataset. 

We first investigate the performance of different detection backbone and framework on HIE dataset, including backbone: ResNet152~\cite{he2016deep}, ResNeXt101~\cite{xie2017aggregated} and SeNet154~\cite{hu2018squeeze}, and different framework: Faster-RCNN~\cite{ren2015faster}, Cascade R-CNN~\cite{cai2018cascade}, and Feature-Pyramid Networks (FPN)~\cite{lin2017feature}. The experimental results on different backbone and methods are given in Table~\ref{tab:det_methods}. The baseline model is Faster RCNN with ResNet50, and we search hyper-parameters on the baseline model then apply to the larger backbone. From table~\ref{tab:det_methods}, we can find that the better backbone (ResNet152 and ResNeXt101) and combining advanced methods (Cascade and FPN) can improve the detection performance, but the SENet154 does not get better performance than ResNet152 even it has superior classification performance on ImageNet. So in our final detection solution, we only adopt ResNet152 and ResNeXt101 as the backbone.

\subsection{The Effects of Extra Data}
In the original train data, there are 764k person bounding boxes in 19 videos with 32.9k frames, and the testing set contains 13 videos with 15.1k frames. Considering the limited number of videos and duplicated image frames, the diversity of train data is not enough. And the train data and test data have many different scenes, thus extra data is crucial for training a superior detection model. Here we investigate the effects of different human detection dataset on HIE, including all the person images in COCO (COCO person, 64k images with 262k boxes)~\cite{lin2014microsoft}, CityPerson (2.9k image with 19k boxes)~\cite{zhang2017citypersons}, CrowndHuman (15k images with 339k boxes)~\cite{shao2018crowdhuman} and self-collected data (2k images with 30k boxes). We investigate the effects on different data based on Faster-RCNN with ResNet50 as the backbone. The experimental results are shown in Table~\ref{tab:det_data}. We can find that the CrowdHuman dataset achieves the largest improvement compared with other datasets, because the CrowdHuman is the most similar scenes with HIE, and both of the two datasets contain plenty of crowded scenes. COCO person contains two times of images than HIE train data, but merging the COCO person does not bring significant improvement and suffer more than three times train time, thus we only merge HIE with CrowdHuman and self-collected data to take a trade-off between detection performance and train time. 

\subsection{Detection in Crowded Scenes}
As there are lots of crowded scenes in HIE2020 dataset, the highly-overlapped instances are hard to detect for the current detection framework. We apply a method aiming to predict instances in crowded scenes~\cite{chu2020detection}, named as ``CrowdDet''. The key idea of CrowdDet is to let each proposal predict a set of correlated instances rather than a single one as the previous detection method. The CrowdDet includes three main contributions for crowded-scenes detection: (1) an EMD loss to minimize the set distance between the two sets of proposals~\cite{tang2014detection}; (3). Set NMS, it will skip normal NMS suppression when two bounding boxes come from the same proposal, which has been proved works in crowded detection; (2). A refine module that takes the combination of predictions and the proposal feature as input, then performs a second round of predicting. We conduct experiments to test the three parts on HIE2020 dataset, and the results are shown in Table~\ref{tab:det_crowded}. Based on the results in the Table, we can find that the three parts do improve the performance in crowded detection. Meanwhile, we apply KD regularization~\cite{yuan2019revisit} in the class's logits of the detection model, which can consistently improve the detection results by 0.5\%-1.4\%.

Finally, based on the above analysis, we train two detection models on HIE by combining extra data with the crowded detection framework: (1). ResNet152 + Cascade RCNN + extra data + emd loss + refine module + set NMS + KD regularization, whose AP is 83.21; (2). ResNeXt101 + Cascade RCNN + extra data + emd loss + refine module + set NMS + KD regularization, whose AP is 83.78;  Then two models are fused with weights 1:1.  

\begin{table}[]
\begin{center}
\fontsize{9pt}{13pt}\selectfont
\caption{Performance comparison (AP and mMR) among different detection backbone and methods on HIE dataset. }
\begin{tabular}{l|c|c}
 \toprule
 Methods or Modules & AP (\%) & MMR (\%) \\
 \midrule
 Baseline (ResNet50 + Faster RCNN) & 61.68 & 74.01 \\
 ResNet152 + Faster RCNN &67.32 & 68.17\\
 ResNet152 + Faster RCNN + FPN &69.77 & 64.83\\
 SENet154 +  Faster RCNN + FPN &65.77 & 68.46\\
 ResNeXt101 + Faster RCNN + FPN &69.53 & 63.91\\
 ResNeXt101 + Cascade RCNN + FPN &71.32 & 61.58\\
 ResNet152 +  Cascade RCNN + FPN &71.06 & 62.55\\
 \bottomrule
\end{tabular}
\label{tab:det_methods}
\end{center}
\end{table}

\begin{table}[]
\begin{center}
\fontsize{9pt}{13pt}\selectfont
\caption{The effects of using extra data for human detection on HIE dataset.}
\begin{tabular}{l|c|c}
 \toprule
 Validation set & AP (\%) & MMR (\%) \\
 \midrule
 HIE data & 61.68 & 74.01 \\
 HIE + COCO person &65.83 & 69.75\\
 HIE + CityPerson &63.71 & 67.43\\
 HIE + CrowdHuman &\textbf{78.22} & \textbf{58.33}\\
 HIE + self-collected data &\textbf{69.39} & \textbf{60.82}\\
 HIE + CrowdHuman + COCO + CityPerson &78.53 & 58.63\\
 \textbf{HIE + CrowdHuman + self-collected data}  &\textbf{81.03} & \textbf{55.58}\\
 HIE + all extra data &81.36 & 55.17\\
 \bottomrule
\end{tabular}
\label{tab:det_data}
\end{center}
\end{table}

\begin{table}[]
\begin{center}
\fontsize{9pt}{13pt}\selectfont
\caption{Detection in Crowded Scenes on HIE dataset.}
\begin{tabular}{l|c|c}
 \toprule
 Validation set & AP (\%) & MMR (\%)\\
 \midrule
 ResNet50 + Faster RCNN + extra data & 81.36 & 55.17 \\
 + emd loss  &81.73 & 53.20\\
 + refine module &81.96 & 50.85\\
 + set NMS &82.05 & 49.63\\
 \bottomrule
\end{tabular}
\label{tab:det_crowded}
\end{center}
\end{table}

\section{Human Tracking}
\label{sec:human_tracking}
We adopted a classic two-stage multi-object tracking scheme composing of detection and box association. After the boxes are generated by the detection phase, DeepSORT~\cite{wojke2017deepsort} is applied to perform the association process and assign the id to the boxes. To improve on the baseline DeepSORT method, we used our own pretrained ReID model to extract deep appearance descriptors. For other parts, we retain a similar structure as described in \cite{wojke2017deepsort}. Besides the improved use of a 
robust ReID feature extractor, we also made modifications in the following two perspectives.

\begin{table}[]
\begin{center}
\fontsize{9pt}{13pt}\selectfont
\caption{Grid search result on the training set of HIE dataset.}
\begin{tabular}{c|c|c|c|c}
 \toprule
 \multicolumn{5}{c}{Parameters}\\
 \midrule
 max-cos-dis & nn-budget & max-age & nn-init & max-iou-dis\\
 0.3 & 256 & 30 & 3 & 0.7\\
 \midrule
 \multicolumn{5}{c}{Performance on training set}\\
 \midrule
 MOTA & MOTP & FP & FN & ID Sw\\
 87.7 & 0.131 & 24839 & 61677 & 10175 \\
 \midrule
 \multicolumn{5}{c}{Performance on testing set}\\
 \midrule
 MOTA & MOTP & FP & FN & ID Sw\\
 61.85 & 0.78 & 3710 & 21211 & 1657 \\
 \bottomrule
\end{tabular}
\label{tab:grid_search_result}
\end{center}
\end{table}

\subsection{Optimal parameter search}
To obtain an optimal set of parameters for the tracking method to work well on the testing set, we performed a grid search on the training set. The search space is designed for five parameters, respectively the maximum cosine distance (max-cos-dis), the number of object appearance history to keep (nn-budget), the max age of a track (max-age), the number of frames for a tentative track to be confirmed (nn-init), and the maximum IoU distance (max-iou-dis). The optimal parameter and results on both training and testing set are shown in Table~\ref{tab:grid_search_result}. The performance in the table is also the performance of the final tracker that we used in pose tracking. 

\subsection{Combination of ReID features}
We tested different strategies for combining ReID features to be used by the DeepSORT structure. (a) Linear combination of original features and features extracted from the horizontal flipped human; (b) Linear combination of ReID features extracted by models trained using different scales. The results on the testing set using a preliminary object detection result can be seen in Table~\ref{tab:tracking_test}. 

\begin{table}[]
\begin{center}
\fontsize{9pt}{13pt}\selectfont
\caption{Detection in Crowded Scenes on HIE dataset.}
\begin{tabular}{l|c|c|c|c}
 \toprule
 Method & MOTA & FP & FN & ID Sw\\
 \midrule
 DeepSORT (Baseline) & 27.11 & 5894 & 42668 & 2220 \\
 + Own ReID \& Det   & 53.46 & \textbf{5622} & 24836 & 1963 \\
 + Horizontal flip   & 53.50 & 5653 & 24798 & 1949 \\
 + Scale             & \textbf{53.96} & 5922 & \textbf{24501} & \textbf{1657} \\
 \bottomrule
\end{tabular}
\label{tab:tracking_test}
\end{center}
\end{table}

\section{Pose estimation}
\label{sec:pose_estimation}
\subsection{Framework}

Two state-of-the-art single-person pose estimation models are used in our method as the baseline, respectively HRNet~\cite{sun2019HRNet} and SimpleNet~\cite{xiao2018SimpleNet}. FPN~\cite{lin2017FPN} is embedded into the structure of SimpleNet so that the performance on the small human bounding boxes can be improved. The results of HRNet and SimpleNet with FPN are then fused by averaging the heatmaps. 

\subsection{Optical Flow Smoothing}

The selected models consider the estimation of poses in different pictures as individual predictions. However, human actions are continuous in videos, which means the predictions in consecutive frames should be close to each other. Therefore, we advantage of the optical flow to improve the smoothness of our pose prediction.

We proposed to smooth the poses predicted in the current frame using the previous as well as next frame using optical flow. Given a human instance with joints coordinates set $J^{k-1}_i$ in the previous frame $I^{k-1}$, the estimated joints coordinates in the current frame $\hat{J}^{k-1\xrightarrow{}k}_i$ can be obtained using the optical field $F_{k-1\xrightarrow{}k}$ between $I^{k-1}$ and $I^k$ by propagating $J^{k-1}_i$ according to $F_{k-1\xrightarrow{}k}$. Specifically, for each joint location $(x, y)$ in $J^{k-1}_i$, the propagated joint location will be $(x + \delta x, y + \delta y$, where $\delta x$, $\delta y$ are the flow field values at joint location $(x, y)$. Similarly, we can estimate the current frame $\hat{J}^{k+1\xrightarrow{}k}_i$ from the next frame in the same way. Finally, we obtain the predicted $J^{k}_i$ as follows:
\begin{equation}
\label{eqn1}
J^{k}_i = \alpha\cdot\hat{J}^{k-1\xrightarrow{}k}_i + \alpha\cdot\hat{J}^{k+1\xrightarrow{}k}_i + (1-2\alpha)\cdot J^k_i,
\end{equation}
where the $J^k_i$ is the prediction of current frame from our pose estimation network and the $\alpha$ is used to weighted average the three terms.

The person identities used in optical flow smoothing is obtained using the tracking method described before. The final pose estimation score is presented in Table.~\ref{tab:pose_estimation_res}.

\begin{table}[]
\begin{center}
\fontsize{9pt}{13pt}\selectfont
\caption{Pose estimation result.}
\begin{tabular}{l|c|c|c}
 \toprule
  & w\_AP @ avg & w\_AP @ 50 & w\_AP @ 75 \\
 \midrule
 Ours & 56.34 &69.60 & 53.02\\
 \bottomrule
\end{tabular}
\label{tab:pose_estimation_res}
\end{center}
\end{table}

\subsection{Pose tracking}
The overall procedure for the pose tracking composes of several stages. (a) DeepSORT with our own ReID feature and private detection method is applied to generate ID for each person. (b) Human pose is individually estimated for each box in the result of human tracking. (c) Optical flow smoothing is performed as described, using the human id generated in (a) and human pose generated in (b). The final pose is obtained using Eq.~\ref{eqn1}. The identity of the pose is automatically transferred from the result of DeepSORT to the final poses with the corresponding bounding boxes. The final pose tracking result is presented in Table.~\ref{tab:final_res}.

\begin{table}[]
\begin{center}
\fontsize{9pt}{13pt}\selectfont
\caption{Pose tracking result.}
\begin{tabular}{l|c|c|c}
 \toprule
      &MOTA   & MOTP  & Total\_AP\\
 \midrule
 Ours & 61.79 & 54.97 & 76.55 \\
 \bottomrule
\end{tabular}
\label{tab:final_res}
\end{center}
\end{table}

\section{Conclusion}
\label{sec:conclusion}
This paper presents a pipeline to solve the pose tracking problem in the video. The problem is decomposed into two separate problems that can be solved individually. First, for the human detection problem in crowded scenes, we investigate the effects of common detection frameworks on HIE2020 dataset, then we add extra data to overcome the overfitting problem and apply one proposal for multiple predictions to relieve the difficulty on detecting highly-overlapping instances. After obtaining human positions, we apply DeepSORT to perform box association and assign ID for the boxes and made improvements on two perspectives, respectively in terms of parameter search and ReID features. Then we apply our effective single-person pose estimation model to generate accurate pose predictions and the predictions are smoothed by the proposed optical flow smoothing algorithm. The identity of the boxes are automatically transferred to the corresponding pose. Overall, the proposed pipeline performs well.

\bibliographystyle{ACM-Reference-Format}
\bibliography{HIE}


\begin{thebibliography}{28}


\ifx \showCODEN    \undefined \def \showCODEN     #1{\unskip}     \fi
\ifx \showDOI      \undefined \def \showDOI       #1{#1}\fi
\ifx \showISBNx    \undefined \def \showISBNx     #1{\unskip}     \fi
\ifx \showISBNxiii \undefined \def \showISBNxiii  #1{\unskip}     \fi
\ifx \showISSN     \undefined \def \showISSN      #1{\unskip}     \fi
\ifx \showLCCN     \undefined \def \showLCCN      #1{\unskip}     \fi
\ifx \shownote     \undefined \def \shownote      #1{#1}          \fi
\ifx \showarticletitle \undefined \def \showarticletitle #1{#1}   \fi
\ifx \showURL      \undefined \def \showURL       {\relax}        \fi
\providecommand\bibfield[2]{#2}
\providecommand\bibinfo[2]{#2}
\providecommand\natexlab[1]{#1}
\providecommand\showeprint[2][]{arXiv:#2}

\bibitem[\protect\citeauthoryear{Cai and Vasconcelos}{Cai and
  Vasconcelos}{2018}]%
        {cai2018cascade}
\bibfield{author}{\bibinfo{person}{Zhaowei Cai} {and} \bibinfo{person}{Nuno
  Vasconcelos}.} \bibinfo{year}{2018}\natexlab{}.
\newblock \showarticletitle{Cascade r-cnn: Delving into high quality object
  detection}. In \bibinfo{booktitle}{\emph{Proceedings of the IEEE conference
  on computer vision and pattern recognition}}. \bibinfo{pages}{6154--6162}.
\newblock


\bibitem[\protect\citeauthoryear{Chu, Zheng, Zhang, and Sun}{Chu
  et~al\mbox{.}}{2020}]%
        {chu2020detection}
\bibfield{author}{\bibinfo{person}{Xuangeng Chu}, \bibinfo{person}{Anlin
  Zheng}, \bibinfo{person}{Xiangyu Zhang}, {and} \bibinfo{person}{Jian Sun}.}
  \bibinfo{year}{2020}\natexlab{}.
\newblock \showarticletitle{Detection in Crowded Scenes: One Proposal, Multiple
  Predictions}. In \bibinfo{booktitle}{\emph{Proceedings of the IEEE/CVF
  Conference on Computer Vision and Pattern Recognition}}.
  \bibinfo{pages}{12214--12223}.
\newblock


\bibitem[\protect\citeauthoryear{Dollar, Wojek, Schiele, and Perona}{Dollar
  et~al\mbox{.}}{2011}]%
        {dollar2011pedestrian}
\bibfield{author}{\bibinfo{person}{Piotr Dollar}, \bibinfo{person}{Christian
  Wojek}, \bibinfo{person}{Bernt Schiele}, {and} \bibinfo{person}{Pietro
  Perona}.} \bibinfo{year}{2011}\natexlab{}.
\newblock \showarticletitle{Pedestrian detection: An evaluation of the state of
  the art}.
\newblock \bibinfo{journal}{\emph{IEEE transactions on pattern analysis and
  machine intelligence}} \bibinfo{volume}{34}, \bibinfo{number}{4}
  (\bibinfo{year}{2011}), \bibinfo{pages}{743--761}.
\newblock


\bibitem[\protect\citeauthoryear{Girshick}{Girshick}{2015}]%
        {girshick2015fast}
\bibfield{author}{\bibinfo{person}{Ross Girshick}.}
  \bibinfo{year}{2015}\natexlab{}.
\newblock \showarticletitle{Fast r-cnn}. In
  \bibinfo{booktitle}{\emph{Proceedings of the IEEE international conference on
  computer vision}}. \bibinfo{pages}{1440--1448}.
\newblock


\bibitem[\protect\citeauthoryear{Girshick, Donahue, Darrell, and
  Malik}{Girshick et~al\mbox{.}}{2014}]%
        {girshick2014rich}
\bibfield{author}{\bibinfo{person}{Ross Girshick}, \bibinfo{person}{Jeff
  Donahue}, \bibinfo{person}{Trevor Darrell}, {and} \bibinfo{person}{Jitendra
  Malik}.} \bibinfo{year}{2014}\natexlab{}.
\newblock \showarticletitle{Rich feature hierarchies for accurate object
  detection and semantic segmentation}. In
  \bibinfo{booktitle}{\emph{Proceedings of the IEEE conference on computer
  vision and pattern recognition}}. \bibinfo{pages}{580--587}.
\newblock


\bibitem[\protect\citeauthoryear{He, Gkioxari, Doll{\'a}r, and Girshick}{He
  et~al\mbox{.}}{2017}]%
        {he2017mask}
\bibfield{author}{\bibinfo{person}{Kaiming He}, \bibinfo{person}{Georgia
  Gkioxari}, \bibinfo{person}{Piotr Doll{\'a}r}, {and} \bibinfo{person}{Ross
  Girshick}.} \bibinfo{year}{2017}\natexlab{}.
\newblock \showarticletitle{Mask r-cnn}. In
  \bibinfo{booktitle}{\emph{Proceedings of the IEEE international conference on
  computer vision}}. \bibinfo{pages}{2961--2969}.
\newblock


\bibitem[\protect\citeauthoryear{He, Zhang, Ren, and Sun}{He
  et~al\mbox{.}}{2016}]%
        {he2016deep}
\bibfield{author}{\bibinfo{person}{Kaiming He}, \bibinfo{person}{Xiangyu
  Zhang}, \bibinfo{person}{Shaoqing Ren}, {and} \bibinfo{person}{Jian Sun}.}
  \bibinfo{year}{2016}\natexlab{}.
\newblock \showarticletitle{Deep residual learning for image recognition}. In
  \bibinfo{booktitle}{\emph{Proceedings of the IEEE conference on computer
  vision and pattern recognition}}. \bibinfo{pages}{770--778}.
\newblock


\bibitem[\protect\citeauthoryear{Hu, Shen, and Sun}{Hu et~al\mbox{.}}{2018}]%
        {hu2018squeeze}
\bibfield{author}{\bibinfo{person}{Jie Hu}, \bibinfo{person}{Li Shen}, {and}
  \bibinfo{person}{Gang Sun}.} \bibinfo{year}{2018}\natexlab{}.
\newblock \showarticletitle{Squeeze-and-excitation networks}. In
  \bibinfo{booktitle}{\emph{Proceedings of the IEEE conference on computer
  vision and pattern recognition}}. \bibinfo{pages}{7132--7141}.
\newblock


\bibitem[\protect\citeauthoryear{Lin, Doll{\'a}r, Girshick, He, Hariharan, and
  Belongie}{Lin et~al\mbox{.}}{2017a}]%
        {lin2017FPN}
\bibfield{author}{\bibinfo{person}{Tsung-Yi Lin}, \bibinfo{person}{Piotr
  Doll{\'a}r}, \bibinfo{person}{Ross Girshick}, \bibinfo{person}{Kaiming He},
  \bibinfo{person}{Bharath Hariharan}, {and} \bibinfo{person}{Serge Belongie}.}
  \bibinfo{year}{2017}\natexlab{a}.
\newblock \showarticletitle{Feature pyramid networks for object detection}. In
  \bibinfo{booktitle}{\emph{Proceedings of the IEEE conference on computer
  vision and pattern recognition}}. \bibinfo{pages}{2117--2125}.
\newblock


\bibitem[\protect\citeauthoryear{Lin, Doll{\'a}r, Girshick, He, Hariharan, and
  Belongie}{Lin et~al\mbox{.}}{2017b}]%
        {lin2017feature}
\bibfield{author}{\bibinfo{person}{Tsung-Yi Lin}, \bibinfo{person}{Piotr
  Doll{\'a}r}, \bibinfo{person}{Ross Girshick}, \bibinfo{person}{Kaiming He},
  \bibinfo{person}{Bharath Hariharan}, {and} \bibinfo{person}{Serge Belongie}.}
  \bibinfo{year}{2017}\natexlab{b}.
\newblock \showarticletitle{Feature pyramid networks for object detection}. In
  \bibinfo{booktitle}{\emph{Proceedings of the IEEE conference on computer
  vision and pattern recognition}}. \bibinfo{pages}{2117--2125}.
\newblock


\bibitem[\protect\citeauthoryear{Lin, Goyal, Girshick, He, and Doll{\'a}r}{Lin
  et~al\mbox{.}}{2017c}]%
        {lin2017focal}
\bibfield{author}{\bibinfo{person}{Tsung-Yi Lin}, \bibinfo{person}{Priya
  Goyal}, \bibinfo{person}{Ross Girshick}, \bibinfo{person}{Kaiming He}, {and}
  \bibinfo{person}{Piotr Doll{\'a}r}.} \bibinfo{year}{2017}\natexlab{c}.
\newblock \showarticletitle{Focal loss for dense object detection}. In
  \bibinfo{booktitle}{\emph{Proceedings of the IEEE international conference on
  computer vision}}. \bibinfo{pages}{2980--2988}.
\newblock


\bibitem[\protect\citeauthoryear{Lin, Maire, Belongie, Hays, Perona, Ramanan,
  Doll{\'a}r, and Zitnick}{Lin et~al\mbox{.}}{2014}]%
        {lin2014microsoft}
\bibfield{author}{\bibinfo{person}{Tsung-Yi Lin}, \bibinfo{person}{Michael
  Maire}, \bibinfo{person}{Serge Belongie}, \bibinfo{person}{James Hays},
  \bibinfo{person}{Pietro Perona}, \bibinfo{person}{Deva Ramanan},
  \bibinfo{person}{Piotr Doll{\'a}r}, {and} \bibinfo{person}{C~Lawrence
  Zitnick}.} \bibinfo{year}{2014}\natexlab{}.
\newblock \showarticletitle{Microsoft coco: Common objects in context}. In
  \bibinfo{booktitle}{\emph{European conference on computer vision}}. Springer,
  \bibinfo{pages}{740--755}.
\newblock


\bibitem[\protect\citeauthoryear{Lin, Liu, Liu, Li, Qi, Qian, Wang, Sebe, Xu,
  Xiong, et~al\mbox{.}}{Lin et~al\mbox{.}}{2020}]%
        {lin2020human}
\bibfield{author}{\bibinfo{person}{Weiyao Lin}, \bibinfo{person}{Huabin Liu},
  \bibinfo{person}{Shizhan Liu}, \bibinfo{person}{Yuxi Li},
  \bibinfo{person}{Guo-Jun Qi}, \bibinfo{person}{Rui Qian},
  \bibinfo{person}{Tao Wang}, \bibinfo{person}{Nicu Sebe},
  \bibinfo{person}{Ning Xu}, \bibinfo{person}{Hongkai Xiong}, {et~al\mbox{.}}}
  \bibinfo{year}{2020}\natexlab{}.
\newblock \showarticletitle{Human in Events: A Large-Scale Benchmark for
  Human-centric Video Analysis in Complex Events}.
\newblock \bibinfo{journal}{\emph{arXiv preprint arXiv:2005.04490}}
  (\bibinfo{year}{2020}).
\newblock


\bibitem[\protect\citeauthoryear{Liu, Anguelov, Erhan, Szegedy, Reed, Fu, and
  Berg}{Liu et~al\mbox{.}}{2016}]%
        {liu2016ssd}
\bibfield{author}{\bibinfo{person}{Wei Liu}, \bibinfo{person}{Dragomir
  Anguelov}, \bibinfo{person}{Dumitru Erhan}, \bibinfo{person}{Christian
  Szegedy}, \bibinfo{person}{Scott Reed}, \bibinfo{person}{Cheng-Yang Fu},
  {and} \bibinfo{person}{Alexander~C Berg}.} \bibinfo{year}{2016}\natexlab{}.
\newblock \showarticletitle{Ssd: Single shot multibox detector}. In
  \bibinfo{booktitle}{\emph{European conference on computer vision}}. Springer,
  \bibinfo{pages}{21--37}.
\newblock


\bibitem[\protect\citeauthoryear{Redmon, Divvala, Girshick, and Farhadi}{Redmon
  et~al\mbox{.}}{2016}]%
        {redmon2016you}
\bibfield{author}{\bibinfo{person}{Joseph Redmon}, \bibinfo{person}{Santosh
  Divvala}, \bibinfo{person}{Ross Girshick}, {and} \bibinfo{person}{Ali
  Farhadi}.} \bibinfo{year}{2016}\natexlab{}.
\newblock \showarticletitle{You only look once: Unified, real-time object
  detection}. In \bibinfo{booktitle}{\emph{Proceedings of the IEEE conference
  on computer vision and pattern recognition}}. \bibinfo{pages}{779--788}.
\newblock


\bibitem[\protect\citeauthoryear{Redmon and Farhadi}{Redmon and
  Farhadi}{2016}]%
        {redmon2016yolo9000}
\bibfield{author}{\bibinfo{person}{Joseph Redmon} {and} \bibinfo{person}{Ali
  Farhadi}.} \bibinfo{year}{2016}\natexlab{}.
\newblock \showarticletitle{YOLO9000: Better, Faster, Stronger}.
\newblock \bibinfo{journal}{\emph{arXiv preprint arXiv:1612.08242}}
  (\bibinfo{year}{2016}).
\newblock


\bibitem[\protect\citeauthoryear{Ren, He, Girshick, and Sun}{Ren
  et~al\mbox{.}}{2015}]%
        {ren2015faster}
\bibfield{author}{\bibinfo{person}{Shaoqing Ren}, \bibinfo{person}{Kaiming He},
  \bibinfo{person}{Ross Girshick}, {and} \bibinfo{person}{Jian Sun}.}
  \bibinfo{year}{2015}\natexlab{}.
\newblock \showarticletitle{Faster r-cnn: Towards real-time object detection
  with region proposal networks}. In \bibinfo{booktitle}{\emph{Advances in
  neural information processing systems}}. \bibinfo{pages}{91--99}.
\newblock


\bibitem[\protect\citeauthoryear{Shao, Zhao, Li, Xiao, Yu, Zhang, and Sun}{Shao
  et~al\mbox{.}}{2018}]%
        {shao2018crowdhuman}
\bibfield{author}{\bibinfo{person}{Shuai Shao}, \bibinfo{person}{Zijian Zhao},
  \bibinfo{person}{Boxun Li}, \bibinfo{person}{Tete Xiao},
  \bibinfo{person}{Gang Yu}, \bibinfo{person}{Xiangyu Zhang}, {and}
  \bibinfo{person}{Jian Sun}.} \bibinfo{year}{2018}\natexlab{}.
\newblock \showarticletitle{Crowdhuman: A benchmark for detecting human in a
  crowd}.
\newblock \bibinfo{journal}{\emph{arXiv preprint arXiv:1805.00123}}
  (\bibinfo{year}{2018}).
\newblock


\bibitem[\protect\citeauthoryear{Sun, Xiao, Liu, and Wang}{Sun
  et~al\mbox{.}}{2019}]%
        {sun2019HRNet}
\bibfield{author}{\bibinfo{person}{Ke Sun}, \bibinfo{person}{Bin Xiao},
  \bibinfo{person}{Dong Liu}, {and} \bibinfo{person}{Jingdong Wang}.}
  \bibinfo{year}{2019}\natexlab{}.
\newblock \showarticletitle{Deep high-resolution representation learning for
  human pose estimation}. In \bibinfo{booktitle}{\emph{Proceedings of the IEEE
  conference on computer vision and pattern recognition}}.
  \bibinfo{pages}{5693--5703}.
\newblock


\bibitem[\protect\citeauthoryear{Tang, Andriluka, and Schiele}{Tang
  et~al\mbox{.}}{2014}]%
        {tang2014detection}
\bibfield{author}{\bibinfo{person}{Siyu Tang}, \bibinfo{person}{Mykhaylo
  Andriluka}, {and} \bibinfo{person}{Bernt Schiele}.}
  \bibinfo{year}{2014}\natexlab{}.
\newblock \showarticletitle{Detection and tracking of occluded people}.
\newblock \bibinfo{journal}{\emph{International Journal of Computer Vision}}
  \bibinfo{volume}{110}, \bibinfo{number}{1} (\bibinfo{year}{2014}),
  \bibinfo{pages}{58--69}.
\newblock


\bibitem[\protect\citeauthoryear{Wang, Yuan, Zhang, and Feng}{Wang
  et~al\mbox{.}}{2019a}]%
        {wang2019distilling}
\bibfield{author}{\bibinfo{person}{Tao Wang}, \bibinfo{person}{Li Yuan},
  \bibinfo{person}{Xiaopeng Zhang}, {and} \bibinfo{person}{Jiashi Feng}.}
  \bibinfo{year}{2019}\natexlab{a}.
\newblock \showarticletitle{Distilling object detectors with fine-grained
  feature imitation}. In \bibinfo{booktitle}{\emph{Proceedings of the IEEE
  Conference on Computer Vision and Pattern Recognition}}.
  \bibinfo{pages}{4933--4942}.
\newblock


\bibitem[\protect\citeauthoryear{Wang, Zhang, Yuan, and Feng}{Wang
  et~al\mbox{.}}{2019b}]%
        {wang2019few}
\bibfield{author}{\bibinfo{person}{Tao Wang}, \bibinfo{person}{Xiaopeng Zhang},
  \bibinfo{person}{Li Yuan}, {and} \bibinfo{person}{Jiashi Feng}.}
  \bibinfo{year}{2019}\natexlab{b}.
\newblock \showarticletitle{Few-shot adaptive faster r-cnn}. In
  \bibinfo{booktitle}{\emph{Proceedings of the IEEE Conference on Computer
  Vision and Pattern Recognition}}. \bibinfo{pages}{7173--7182}.
\newblock


\bibitem[\protect\citeauthoryear{Wojke, Bewley, and Paulus}{Wojke
  et~al\mbox{.}}{2017}]%
        {wojke2017deepsort}
\bibfield{author}{\bibinfo{person}{Nicolai Wojke}, \bibinfo{person}{Alex
  Bewley}, {and} \bibinfo{person}{Dietrich Paulus}.}
  \bibinfo{year}{2017}\natexlab{}.
\newblock \showarticletitle{Simple online and realtime tracking with a deep
  association metric}. In \bibinfo{booktitle}{\emph{2017 IEEE international
  conference on image processing (ICIP)}}. IEEE, \bibinfo{pages}{3645--3649}.
\newblock


\bibitem[\protect\citeauthoryear{Xiao, Wu, and Wei}{Xiao et~al\mbox{.}}{2018}]%
        {xiao2018SimpleNet}
\bibfield{author}{\bibinfo{person}{Bin Xiao}, \bibinfo{person}{Haiping Wu},
  {and} \bibinfo{person}{Yichen Wei}.} \bibinfo{year}{2018}\natexlab{}.
\newblock \showarticletitle{Simple baselines for human pose estimation and
  tracking}. In \bibinfo{booktitle}{\emph{Proceedings of the European
  conference on computer vision (ECCV)}}. \bibinfo{pages}{466--481}.
\newblock


\bibitem[\protect\citeauthoryear{Xie, Girshick, Doll{\'a}r, Tu, and He}{Xie
  et~al\mbox{.}}{2017}]%
        {xie2017aggregated}
\bibfield{author}{\bibinfo{person}{Saining Xie}, \bibinfo{person}{Ross
  Girshick}, \bibinfo{person}{Piotr Doll{\'a}r}, \bibinfo{person}{Zhuowen Tu},
  {and} \bibinfo{person}{Kaiming He}.} \bibinfo{year}{2017}\natexlab{}.
\newblock \showarticletitle{Aggregated residual transformations for deep neural
  networks}. In \bibinfo{booktitle}{\emph{Proceedings of the IEEE conference on
  computer vision and pattern recognition}}. \bibinfo{pages}{1492--1500}.
\newblock


\bibitem[\protect\citeauthoryear{Yuan, Tay, Li, Wang, and Feng}{Yuan
  et~al\mbox{.}}{2019}]%
        {yuan2019revisit}
\bibfield{author}{\bibinfo{person}{Li Yuan}, \bibinfo{person}{Francis~EH Tay},
  \bibinfo{person}{Guilin Li}, \bibinfo{person}{Tao Wang}, {and}
  \bibinfo{person}{Jiashi Feng}.} \bibinfo{year}{2019}\natexlab{}.
\newblock \showarticletitle{Revisit knowledge distillation: a teacher-free
  framework}.
\newblock \bibinfo{journal}{\emph{arXiv preprint arXiv:1909.11723}}
  (\bibinfo{year}{2019}).
\newblock


\bibitem[\protect\citeauthoryear{Zhang, Benenson, and Schiele}{Zhang
  et~al\mbox{.}}{2017}]%
        {zhang2017citypersons}
\bibfield{author}{\bibinfo{person}{Shanshan Zhang}, \bibinfo{person}{Rodrigo
  Benenson}, {and} \bibinfo{person}{Bernt Schiele}.}
  \bibinfo{year}{2017}\natexlab{}.
\newblock \showarticletitle{Citypersons: A diverse dataset for pedestrian
  detection}. In \bibinfo{booktitle}{\emph{Proceedings of the IEEE Conference
  on Computer Vision and Pattern Recognition}}. \bibinfo{pages}{3213--3221}.
\newblock


\bibitem[\protect\citeauthoryear{Zhou, Zhao, Li, Yuan, and Feng}{Zhou
  et~al\mbox{.}}{2018}]%
        {zhou2018object}
\bibfield{author}{\bibinfo{person}{Li Zhou}, \bibinfo{person}{Jian Zhao},
  \bibinfo{person}{Jianshu Li}, \bibinfo{person}{Li Yuan}, {and}
  \bibinfo{person}{Jiashi Feng}.} \bibinfo{year}{2018}\natexlab{}.
\newblock \showarticletitle{Object Relation Detection Based on One-shot
  Learning}.
\newblock \bibinfo{journal}{\emph{arXiv preprint arXiv:1807.05857}}
  (\bibinfo{year}{2018}).
\newblock


\end{thebibliography}

\appendix
\end{document}